\documentclass[letterpaper]{article} 
\usepackage{aaai2026}  

\usepackage{pifont}
\usepackage{times}  
\usepackage{helvet}  
\usepackage{courier}  
\usepackage[hyphens]{url}  
\usepackage{graphicx} 
\usepackage{natbib}  
\usepackage{caption} 
\usepackage{booktabs}
\usepackage{bm}
\usepackage{pifont}
\usepackage{colortbl}
\usepackage{amsmath}
\usepackage{amssymb}
\usepackage{amsfonts}

\definecolor{headerbg}{gray}{0.90}
\definecolor{rowbg}{gray}{0.95}
\usepackage{algorithm}
\usepackage{algorithmic}
\usepackage{multirow}   
\usepackage{siunitx}    

\usepackage{newfloat}
\usepackage{listings}
\DeclareCaptionStyle{ruled}{labelfont=normalfont,labelsep=colon,strut=off} 
\floatstyle{ruled}
\newfloat{listing}{tb}{lst}{}
\floatname{listing}{Listing}
\title{DeepGB-TB: A Risk-Balanced Cross-Attention Gradient-Boosted Convolutional Network for Rapid, Interpretable Tuberculosis Screening}
\author{
    Zhixiang Lu\textsuperscript{\rm 1,2}\equalcontrib,
    Yulong Li\textsuperscript{\rm 1,2,3}\equalcontrib, Feilong Tang\textsuperscript{\rm 3}, Zhengyong Jiang\textsuperscript{\rm 1}, Chong Li\textsuperscript{\rm 1}, Mian Zhou\textsuperscript{\rm 1}\textsuperscript{\textdagger}, Tenglong Li\textsuperscript{\rm 4}\textsuperscript{\textdagger}, Jionglong Su\textsuperscript{\rm 1}\thanks{Corresponding authors.}
}
\affiliations{
    \textsuperscript{\rm 1}School of Artificial Intelligence and Advanced Computing, Xi’an Jiaotong-Liverpool University\\
    \textsuperscript{\rm 2}Department of Computer Science, University of Liverpool\\
    \textsuperscript{\rm 3}Mohamed bin Zayed University of Artificial Intelligence\\
    \textsuperscript{\rm 4}Wisdom Lake Academy of Pharmacy, Xi’an Jiaotong-Liverpool University

    \{Mian.Zhou, Tenglong.Li, Jionglong.Su\}@xjtlu.edu.cn
}

\usepackage{bibentry}

\begin{document}

\maketitle
\begin{abstract}
Large-scale tuberculosis (TB) screening is limited by the high cost and operational complexity of traditional diagnostics, creating a need for artificial-intelligence solutions. We propose DeepGB-TB, a non-invasive system that instantly assigns TB risk scores using only cough audio and basic demographic data. The model couples a lightweight one-dimensional convolutional neural network for audio processing with a gradient-boosted decision tree for tabular features. Its principal innovation is a Cross-Modal Bidirectional Cross-Attention module (CM-BCA) that iteratively exchanges salient cues between modalities, emulating the way clinicians integrate symptoms and risk factors. To meet the clinical priority of minimizing missed cases, we design a Tuberculosis Risk-Balanced Loss (TRBL) that places stronger penalties on false-negative predictions, thereby reducing high-risk misclassifications. DeepGB-TB is evaluated on a diverse dataset of 1,105 patients collected across seven countries, achieving an AUROC of 0.903 and an F1-score of 0.851, representing a new state of the art. Its computational efficiency enables real-time, offline inference directly on common mobile devices, making it ideal for low-resource settings. Importantly, the system produces clinically validated explanations that promote trust and adoption by frontline health workers. By coupling AI innovation with public-health requirements for speed, affordability, and reliability, DeepGB-TB offers a tool for advancing global TB control.
\end{abstract}

\section{Introduction}
Tuberculosis (TB) inflicts a staggering toll on global health, remaining a leading infectious cause of death worldwide~\citep{who2023, who2024}. The cornerstone of TB control, early diagnosis and treatment, is severely undermined in many regions by the limitations of conventional methods. Diagnostic tools like sputum smear microscopy and nucleic acid amplification tests (NAATs) either suffer from low sensitivity or are prohibitively expensive and require centralized laboratories and skilled technicians, creating significant barriers to access in low-resource settings~\citep{lawn2011, boehme2010}. This diagnostic gap leads to delayed treatment, increased transmission, and preventable mortality~\citep{who2015}, underscoring an urgent need for accessible, affordable, and scalable screening solutions.

The ubiquity of mobile phones presents a unique opportunity to bridge this gap. Cough, a cardinal symptom of pulmonary TB, contains a wealth of acoustic information that, if properly analyzed, could serve as a non-invasive digital biomarker~\citep{imran2019}. While prior research has explored AI for cough-based analysis, existing models often face critical limitations. Many either rely solely on audio, ignoring crucial demographic and clinical risk factors (e.g., age, sex, exposure history), or they struggle to effectively integrate these heterogeneous data types. Simple concatenation or late-stage fusion of features often fails to capture the complex, non-linear interplay between a patient's background risk and their real-time acoustic symptoms, a process central to a clinician's diagnostic reasoning~\citep{ramachandram2017}.

To address these shortcomings, we propose DeepGB-TB, a novel, multimodal deep learning system designed for end-to-end, instantaneous TB risk stratification. Our architecture is explicitly designed to model the synergy between who the patient is and how they cough. The framework processes two parallel data streams: a lightweight 1D Convolutional Neural Network (CNN) extracts discriminative features from raw cough audio, while demographic data is handled by our first key innovation. (1) The \textbf{Cross-Validated Probability Embedding Module (CVPEM)} transforms raw tabular data into a robust, high-dimensional feature vector, a technique designed to mitigate overfitting and enhance generalization. These distinct data pathways are then unified by (2) the \textbf{Integrated Multimodal Diagnostic Module (IMDM)}. The centerpiece of this module is (3) the \textbf{Cross-Modal Bidirectional Cross-Attention (CM-BCA)} mechanism. This module moves beyond feature fusion and emulates clinical ratiocination by allowing the audio and tabular embeddings to iteratively query each other, mutually refining their representations to focus on the most salient diagnostic indicators.

Furthermore, recognizing that failing to detect a true TB case (a false negative) has far more severe consequences than a false alarm, we introduce a Tuberculosis Risk-Balanced Loss (TRBL) function. This clinically-aligned loss function imposes a greater penalty on false negatives, systematically steering the model towards the high sensitivity required for an effective screening tool.

Our main contributions are summarized as follows:

\text{1)} \textbf{A Novel Hybrid Architecture:} We introduce DeepGB-TB, a carefully designed deep learning framework that synergistically combines a 1D-CNN for audio analysis with a novel CVPEM-enhanced gradient boosting model for tabular data, enabling a holistic patient assessment.

\text{2)} \textbf{Advanced Multimodal Fusion:} We propose the CM-BCA cross-attention module that achieves integration of heterogeneous data by modeling the bidirectional dependencies between acoustic and demographic features.

\text{3)} \textbf{Clinically-Informed Optimization:} We design the TRBL function, a custom loss function that addresses the clinical priority of minimizing false negatives in TB screening, enhancing the model's real-world utility and safety.

\text{4)} \textbf{State-of-the-Art Performance and Deployability:} We demonstrate that DeepGB-TB achieves state-of-the-art results on multi-national dataset and is computationally efficient for real-time, offline deployment on mobile devices, paving the way for equitable access to TB screening.

\section{Related Works}
\label{sec:related_works}

Early work in AI-driven TB diagnosis predominantly focused on analyzing Chest X-rays (CXRs), where Convolutional Neural Networks (CNNs) have achieved radiologist-level performance~\citep{lakhani2017}. However, the reliance on specialized hardware for CXR imaging limits its applicability for widespread, community-level screening. This has motivated a research shift towards more accessible biomarkers, with cough sounds emerging as a leading alternative. The analysis of cough sounds via AI has emerged as a promising, low-cost screening paradigm. Initial approaches relied on traditional audio features paired with classical machine learning models like Logistic Regression~\citep{Cox1958}. More recently, deep learning has become the standard, with architectures such as 1D-CNNs~\citep{Kiranyaz2016} and deeper models like ResNet~\citep{He2016} applied to spectrograms. The state-of-the-art (SOTA) includes pre-trained audio foundation models like Google's HeAR~\citep{hear2024}, which demonstrated superior performance on 33 health-related acoustic tasks, including TB recognition. Nevertheless, HeAR is not open-source and accessible only via a restricted API for online inference, limiting adoption in resource-constrained settings and creating a critical gap for an efficient, open-source TB screening tool. Clinical TB diagnosis is inherently multimodal, integrating symptoms, history, and risk factors. Common multimodal fusion baselines, such as late-fusion CNN-LightGBM ensembles~\citep{Lu2023}, often fail to capture deep inter-modal synergies~\citep{li2025rhythm}. While advanced tabular models like TabTransformer~\citep{huang2020tabtransformer} exist, their effective integration remains challenging. Frontier models include Large Multimodal Models (LMMs) like Qwen-Omni~\citep{Qwen2.5-Omni}, offering general-purpose understanding~\citep{li2025genesis, LiAVSS}. However, LMMs are computationally prohibitive, and their superiority over specialized models for this diagnostic task is unestablished. Furthermore, the poor interpretability of many deep learning models~\citep{multimodal_review, prev_best} remains an obstacle to reliable clinical deployment.

\section{Methodology}

\begin{figure*}
\centering
\includegraphics[width=0.95\textwidth]{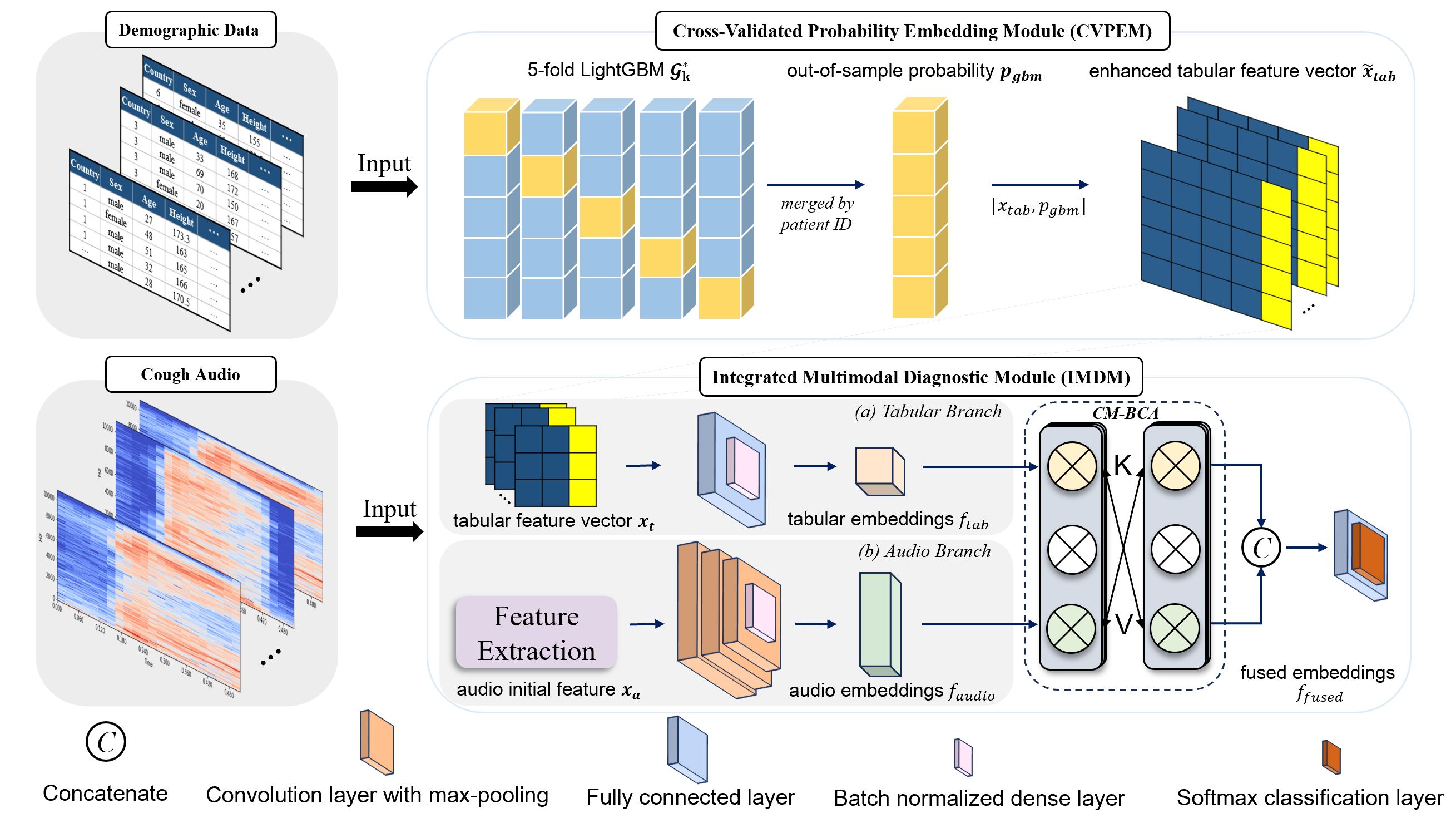}
\caption{The architecture of DeepGB-TB.}
\label{fig1}
\end{figure*}
\subsection{Dataset}

The dataset used in this study~\citep{jaganath2024} was derived from a retrospective, multicenter case-control investigation encompassing 1,105 adult participants across seven countries (India, the Philippines, South Africa, Uganda, Vietnam, Tanzania, and Madagascar). Each participant presented with a new or worsening cough lasting at least two weeks. Comprehensive clinical, demographic, and cough acoustic data were collected for each participant. Participants were classified as TB positive if testing positive on Mycobacterial culture~\citep{kent1985}, Xpert MTB/RIF~\citep{boehme2010}, or Xpert Ultra~\citep{chakravorty2017}, and TB negative if negative on all assays. The study received ethical approval from all participating institutional review boards. Exploratory analysis revealed significant group differences: TB prevalence was higher in males (33.2\%) than females (19.7\%). Furthermore, hemoptysis (43.2\% vs.\ 24.3\%), night sweats (39.0\% vs.\ 17.3\%), fever (40.0\% vs.\ 16.1\%), and weight loss (36.5\% vs.\ 11.9\%) were all strongly associated with TB.

\subsection{Data Pre-processing and Feature Extraction}
\label{subsec:preprocessing}
Raw audio recordings first undergo standard pre-processing, including spectral subtraction for noise removal~\citep{ref_noise1} and peak volume normalization. From the cleaned audio segments, we extract a comprehensive feature set. The primary features are Mel-frequency cepstral coefficients (MFCCs)~\citep{davis1980}, which robustly capture spectro-temporal characteristics. The $n$-th coefficient $c_n$ is computed as:
\begin{equation}
    c_n = \sum_{k=1}^{M} (\log S_k) \cos\left[n\left(k - \frac{1}{2}\right)\frac{\pi}{M}\right], \quad n=1, \dots, L
\end{equation}
where $S_k$ are the log-energies from $M$ Mel-scaled filterbanks and $L$ is the number of coefficients. To create a richer acoustic representation, we supplement MFCCs with auxiliary features including spectral centroid, chroma, zero-crossing rate (ZCR), and fundamental frequency (F0)~\citep{giannakopoulos2015}.

\subsection{Statistical Analysis}
\label{subsec:statistics}
To provide an interpretable analysis of the high-dimensional and often collinear acoustic features~\citep{stat_method2, stat_method3}, we employ a classical statistical pipeline. First, we use t-distributed Stochastic Neighbor Embedding (t-SNE)~\citep{tsne2008} for dimensionality reduction, which minimizes the Kullback-Leibler (KL) divergence between the joint probability distributions of the original data, $p_{ij}$, and the low-dimensional embeddings, $q_{ij}$:
\begin{equation}
    C = \sum_i \sum_j p_{ij} \log \frac{p_{ij}}{q_{ij}}
\end{equation}
A logistic regression model is then trained on the resulting embeddings to assess predictive power. We validate the structural integrity of this non-linear mapping using a Mantel correlation test~\citep{mantel1967, tsne_interpret} and confirm the significance of the most discriminative features using Wald and independent two-sample t-tests.

\subsection{Proposed Framework}
The architecture of DeepGB-TB, shown in Figure~\ref{fig1}, integrates two key modules: the Cross-Validated Probability Embedding Module (CVPEM) and the Integrated Multimodal Diagnostic Module (IMDM). CVPEM employs LightGBM to generate robust probability embeddings from tabular demographic data, while IMDM processes cough audio signals via a 1D-CNN to extract temporal features. DeepGB-TB effectively fuses these complementary modalities, offering a comprehensive, robust, and accurate framework for TB diagnosis.

\textbf{Cross-Validated Probability Embedding Module (CVPEM).} In our proposed pipeline, LightGBM is trained with a 5-fold cross-validation scheme on demographic data to generate a stable, out-of-sample probability estimate for each patient. These probabilities are then merged by patient ID and embedded as an additional feature, effectively capturing cross-validated insights into TB risk. Let 
\(\{(x_{\text{tab}, i},\,y_i)\}_{i=1}^n\)
denote a dataset with tabular demographic features 
\(x_{\text{tab}, i} \in \mathbb{R}^{d_{\text{tab}}}\)
and binary labels 
\(y_i \in \{0,1\}\).
We partition the index set 
\(i\in\{1,2,\ldots,n\}\)
into \(K=5\) disjoint subsets 
\(\{\mathcal{I}_1, \mathcal{I}_2, \dots, \mathcal{I}_5\}\)
for 5-fold cross-validation. For each fold \(k \in \{1,\dots,5\}\) to train a LightGBM model \(\mathcal{G}_k^*\) on \(\bigcup_{j \neq k} \mathcal{I}_j\). For each \(i \in \mathcal{I}_k\), compute the out-of-sample probability:
    \begin{equation}
        p_{\text{gbm}, i} = \mathcal{G}_k^*\bigl(x_{\text{tab}, i}\bigr).
    \end{equation}

Collecting these out-of-sample predictions 
\(\{p_{\text{gbm}, i}\}_{i=1}^n\)
yields 
\(p_{\text{gbm}} \in \mathbb{R}^n\).
Each instance \(i\) is thus assigned an enhanced tabular feature vector:
\begin{equation}
    \tilde{x}_{\text{tab}, i} 
    = \bigl[x_{\text{tab}, i},\; p_{\text{gbm}, i}\bigr]
    \;\in\; \mathbb{R}^{d_{\text{tab}} + 1}.
    \label{eq:enhanced_tab}
\end{equation}

By embedding the cross-validated probability \(p_{\text{gbm}, i}\) into the tabular features, CVPEM leverages robust out-of-sample estimates from LightGBM, which are subsequently fed into the downstream network to provide an enriched input representation that boosts both predictive performance and interpretability.

\textbf{Integrated Multimodal Diagnostic Module (IMDM).} Our proposed DeepGB-TB framework incorporates the IMDM to seamlessly combine heterogeneous data modalities for TB diagnosis. In this module, tabular (demographic) data are processed using a boosting-based method (LightGBM) to yield preliminary diagnostic probabilities, while cough audio features are extracted via a 1D-CNN. These two branches are then fused in a unified architecture, effectively bridging classical statistical methods with advanced deep learning techniques to achieve both high predictive performance and improved interpretability. Let 
\(\bigl\{(T_i,\, s_i)\bigr\}_{i=1}^n\)
denote a dataset where 
\(T_i \in \mathbb{R}^{d_{\text{tab}} + 1}\)
is the enhanced tabular feature vector (demographic data plus the cross-validated probability embedding \(p_{\text{gbm}}\)), and 
\(s_i \in \mathbb{R}^L\)
is the raw 1D cough audio signal for the \(i\)-th patient. The IMDM consists of two primary branches (tabular and audio) that converge in a final fusion layer to produce the TB diagnosis probability.

\begin{figure}[t]
\centering
\includegraphics[width=0.9\columnwidth]{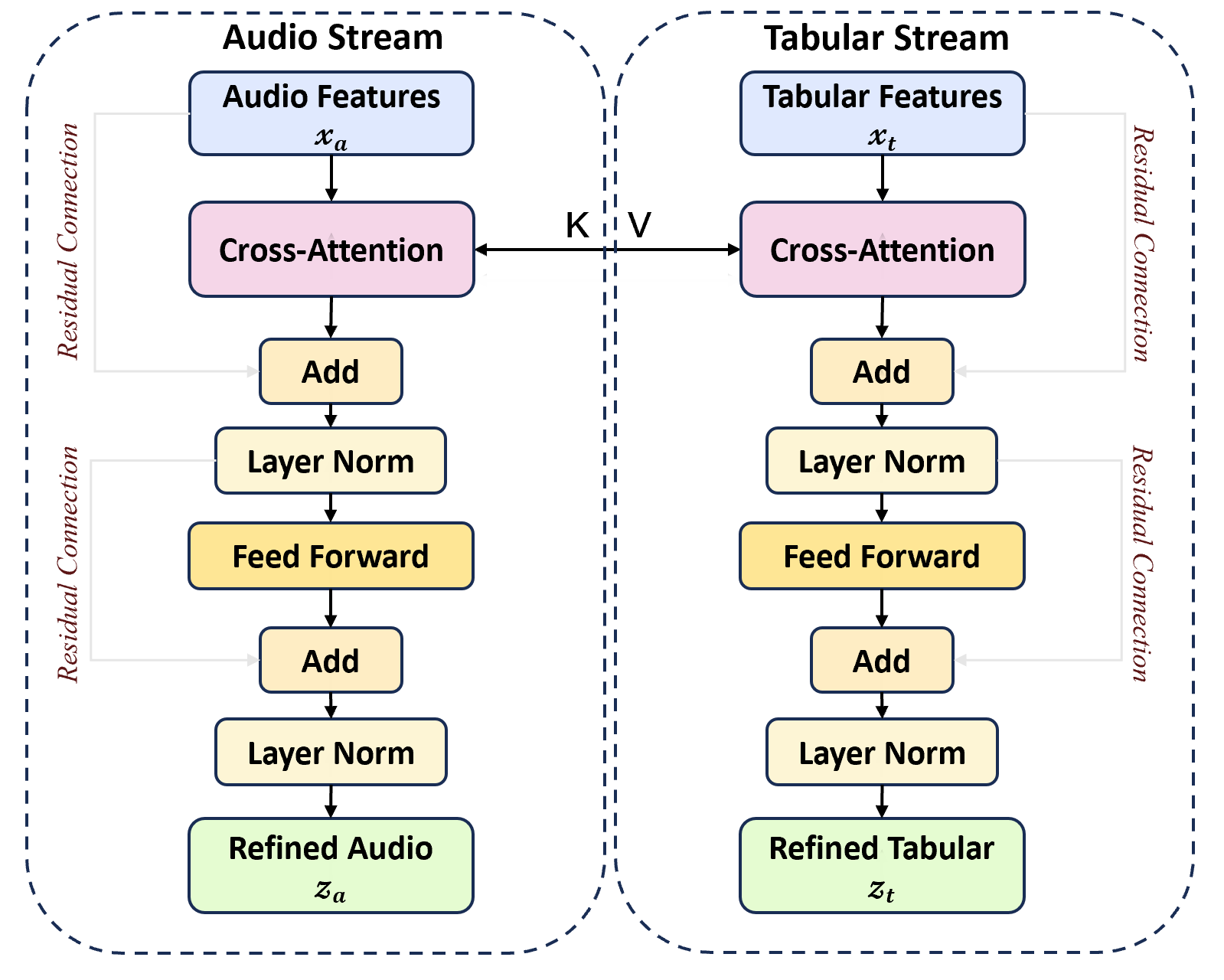} 
\caption{The process of CM-BCA.}
\label{fig:CMBCA}
\end{figure}

\noindent\textbf{(1) Tabular Branch.}
A batch normalization (BN) is applied to stabilize tabular inputs, followed by a dense layer:
\begin{equation}
    T'_i = \text{BN}\bigl(T_i\bigr), 
    \quad
    f_{\text{tab},i} = \text{ReLU}\bigl(W^{(t)}\,T'_i + b^{(t)}\bigr).
    \label{eq:tab_branch}
\end{equation}
\textbf{(2) Audio Branch.}
A preprocessing function (e.g., MFCC extraction) transforms \(s_i\) into an initial feature map:
\begin{equation}
    F^{(0)}_i = \text{Preprocess}\bigl(s_i\bigr).
    \label{eq:preprocess_audio}
\end{equation}
Next, three consecutive convolutional blocks are applied (each comprising a 1D convolution, batch normalization, ReLU activation, and max pooling), yielding:
\begin{equation}
    F^{(l)}_i 
    = \text{MaxPool}\Bigl(\text{ReLU}\bigl(\text{BN}\bigl(\text{Conv1D}_{k_l}\bigl(F^{(l-1)}_i\bigr)\bigr)\bigr)\Bigr),
    \label{eq:conv_blocks}
\end{equation}
The final feature map is flattened and passed through a dense layer with optional dropout:
\begin{equation}
    f_{a,i} = \text{Flatten}\bigl(F^{(3)}_i\bigr),
    \quad
\end{equation}
\begin{equation}
    f_{\text{audio},i} 
    = \text{Dropout}\Bigl(\text{ReLU}\bigl(W^{(a)}\,f_{a,i} + b^{(a)}\bigr)\Bigr)
    \label{eq:audio_branch}
\end{equation}
\textbf{(3) Fusion and Classification.}
The tabular embedding \(f_{\text{tab},i}\) and audio embedding \(f_{\text{audio},i}\) are concatenated:
\begin{equation}
    f_{\text{fused},i} 
    = \text{Concat}\bigl(f_{\text{tab},i},\, f_{\text{audio},i}\bigr).
    \label{eq:fusion}
\end{equation}
A fully connected layer then produces logits \(z_i\), which are mapped to a probability distribution \(\hat{y}_i\) via the SoftMax function:
\begin{equation}
    z_i = W^{(fc)}\,f_{\text{fused},i} + b^{(fc)},
    \quad
    \hat{y}_i = \text{SoftMax}\bigl(z_i\bigr),
    \label{eq:softmax}
\end{equation}
where \(\hat{y}_i \in \mathbb{R}^2\) represents the predicted probabilities for TB-positive. The IMDM thus integrates tabular and audio features in a unified architecture, leveraging the strengths of boosting-based models for tabular data and CNNs for audio analysis, culminating in a robust, multimodal diagnostic framework. For a fair comparison, all models undergo identical training, validation, and test splits.  The training loss function, cross entropy loss~\cite{crossentropy}, was maintained consistently throughout the experiments in deep learning models and was evaluated using metrics such as accuracy, sensitivity (TPR), specificity (TNR), and Area Under the Receiver Operating Characteristic Curve (AUROC) \cite{Fawcett2006}. Statistical tests and a stepwise backward elimination approach \cite{Babyak2001} were employed to assess feature importance.

\begin{algorithm}[t!]
\caption{Cross-Modal Bidirectional Cross-Attention}
\label{alg:cm-bca}
\textbf{Input}: Audio feature $\mathbf{A} \in \mathbb{R}^{d}$, Tabular feature $\mathbf{T} \in \mathbb{R}^{d}$ \\
\textbf{Parameter}: Hidden dimension $d$, number of heads $h$, dropout $p$, max iterations $T$ \\
\textbf{Output}: Updated feature $\mathbf{A}'$, $\mathbf{T}'$
\begin{algorithmic}[1]

\STATE \begin{tabular}[t]{@{}l}
    Let $\mathbf{A}^{(0)} \leftarrow \text{ExpandDims}(\mathbf{A})$, \\
    \qquad $\mathbf{T}^{(0)} \leftarrow \text{ExpandDims}(\mathbf{T})$.
    \end{tabular}
\STATE Let $t \gets 0$.

\WHILE{$t < T$}
    \STATE \texttt{// Tabular-to-Audio Attention}
    \STATE $\mathbf{T}_{att}^{(t)} \gets \text{MHA}(\mathbf{T}^{(t)}, \mathbf{A}^{(t)}, \mathbf{A}^{(t)})$
    \STATE $\mathbf{T}_{int}^{(t)} \gets \text{LayerNorm}(\mathbf{T}^{(t)} + \mathbf{T}_{att}^{(t)})$
    \STATE $\mathbf{T}^{(t+1)} \gets \text{LayerNorm}\big(\mathbf{T}_{int}^{(t)} + \text{FFN}_T(\mathbf{T}_{int}^{(t)})\big)$

    \STATE \texttt{// Audio-to-Tabular Attention}
    \STATE $\mathbf{A}_{att}^{(t)} \gets \text{MHA}(\mathbf{A}^{(t)}, \mathbf{T}^{(t)}, \mathbf{T}^{(t)})$
    \STATE $\mathbf{A}_{int}^{(t)} \gets \text{LayerNorm}(\mathbf{A}^{(t)} + \mathbf{A}_{att}^{(t)})$
    \STATE $\mathbf{A}^{(t+1)} \gets \text{LayerNorm}\big(\mathbf{A}_{int}^{(t)} + \text{FFN}_A(\mathbf{A}_{int}^{(t)})\big)$

    \IF{$\|\mathbf{A}^{(t+1)} - \mathbf{A}^{(t)}\|_2 < \epsilon \;\land\;
         \|\mathbf{T}^{(t+1)} - \mathbf{T}^{(t)}\|_2 < \epsilon$}
        \STATE \textbf{break}
    \ENDIF
    \STATE $t \gets t+1$
\ENDWHILE

\STATE \textbf{return} $\mathbf{A}' \gets \text{Squeeze}(\mathbf{A}^{(t)}), \quad
                        \mathbf{T}' \gets \text{Squeeze}(\mathbf{T}^{(t)})$
\end{algorithmic}
\end{algorithm}

\textbf{Cross-Modal Bidirectional Cross-Attention (CM-BCA).} The CM-BCA is constructed as an operator $\mathcal{F}: \mathbb{R}^d \times \mathbb{R}^d \to \mathbb{R}^d \times \mathbb{R}^d$, which maps a pair of unimodal feature vectors $(\bm{x}_a, \bm{x}_t)$ to a pair of refined, contextually-aware representations $(\bm{z}_a, \bm{z}_t)$. The construction relies on the composition of several fundamental operators, which we define first. For the application of sequence-based operators, each vector $\bm{x} \in \mathbb{R}^d$ is lifted to a singleton sequence $\mathbf{X} \in \mathbb{R}^{1 \times d}$.

\noindent\textbf{(1) Multi-Head Attention Operator.} Let $h$ be the number of heads and $d_k = d/h$. The operator $\mathcal{M}: (\mathbb{R}^{n \times d})^3 \to \mathbb{R}^{n \times d}$ is defined as:
    \begin{equation}
    \mathcal{M}(Q, K, V) := \left( \bigoplus_{i=1}^{h} \sigma_{\text{attn}}(QW_i^Q, KW_i^K, VW_i^V) \right) W^O
    \end{equation}
    where $\bigoplus$ denotes concatenation, $W_i^Q, W_i^K, W_i^V \in \mathbb{R}^{d \times d_k}$ and $W^O \in \mathbb{R}^{hd_k \times d}$ are learnable linear projection matrices, and $\sigma_{\text{attn}}$ is the scaled dot-product attention function:
    \begin{equation}
    \sigma_{\text{attn}}(Q_i, K_i, V_i) := \text{softmax}\left( \frac{Q_i K_i^\top}{\sqrt{d_k}} \right) V_i.
    \end{equation}

\noindent\textbf{(2) Feed-Forward Operator.} The position-wise feed-forward operator $\mathcal{N}: \mathbb{R}^{n \times d} \to \mathbb{R}^{n \times d}$ is a two-layer perceptron:
    \begin{equation}
    \mathcal{N}(\mathbf{X}) := \text{ReLU}(\mathbf{X}W_1 + \bm{b}_1)W_2 + \bm{b}_2.
    \end{equation}

\noindent\textbf{(3) Layer Normalization Operator.} The operator $\mathcal{L}: \mathbb{R}^{n \times d} \to \mathbb{R}^{n \times d}$ normalizes each element $\bm{x} \in \mathbb{R}^d$ as:
\begin{equation}
\mathcal{L}(\bm{x}) := \frac{\bm{x} - \mu\mathbf{1}}{\sqrt{\sigma^2 + \epsilon}}\odot\bm{\gamma} + \bm{\beta},
\end{equation}
where $\mu$ and $\sigma^2$ are the mean and variance of the elements of $\bm{x}$, and $\bm{\gamma}, \bm{\beta} \in \mathbb{R}^d$ are learnable affine parameters.
 
The CM-BCA operator is constructed via the parallel application of two symmetric transformation blocks, $\mathcal{T}_{t \leftarrow a}$ and $\mathcal{T}_{a \leftarrow t}$, which compose the operators defined above.

\noindent\textbf{(4) Unimodal Refinement Block.} The transformation $\mathcal{T}_{t \leftarrow a}$ that refines a target modality $\mathbf{X}_t$ using a source modality $\mathbf{X}_a$ is given by the operator sequence:
\begin{align}
    \mathbf{X}'_t &= \mathcal{L}_{t,1}\left(\mathbf{X}_t + \mathcal{M}_{t \leftarrow a}(\mathbf{X}_t, \mathbf{X}_a, \mathbf{X}_a)\right) \\
    \mathbf{Z}_t &= \mathcal{L}_{t,2}\left(\mathbf{X}'_t + \mathcal{N}_t(\mathbf{X}'_t)\right)
\end{align}
A symmetric transformation $\mathcal{T}_{a \leftarrow t}$ is defined analogously for refining $\mathbf{X}_a$ using $\mathbf{X}_t$.

The complete operator $\mathcal{F}$ is the parallel execution of these blocks on the lifted inputs, followed by a projection $\pi: \mathbb{R}^{1 \times d} \to \mathbb{R}^d$ that removes the sequence dimension:
\begin{equation}
\mathcal{F}(\mathbf{x}_a, \mathbf{x}_t) := \left( \pi\left(\mathcal{T}_{a \leftarrow t}(\mathbf{X}_a, \mathbf{X}_t)\right), \pi\left(\mathcal{T}_{t \leftarrow a}(\mathbf{X}_t, \mathbf{X}_a)\right) \right)
\end{equation}
where $\mathbf{X}_a = \iota(\mathbf{x}_a)$, $\mathbf{X}_t = \iota(\mathbf{x}_t)$, and $\iota: \mathbb{R}^d \to \mathbb{R}^{1 \times d}$ is the initial lifting map. Note that stochastic elements such as Dropout are omitted from this deterministic formulation of the forward map.

\textbf{Tuberculosis Risk-Balanced Loss (TRBL).} To reduce the risk of missed TB cases in high-stakes clinical settings, we introduce the TRBL. This loss function emphasizes false-negative samples by reweighting the standard binary cross-entropy loss, aligning with the priority of maximizing recall in TB detection.

Let $y \in \{0,1\}$ denote the ground-truth label, where $y = 1$ indicates TB-positive. Let $\hat{y} \in (0,1)$ be the predicted probability, and $\lambda > 1$ be a scalar penalty applied to false negatives. Then the TRBL for a single sample is defined as:
\begin{equation}
\mathcal{L}_{\text{BCE}}(y, \hat{y}) = - \left[ y \log(\hat{y}) + (1 - y) \log(1 - \hat{y}) \right]
\label{eq:bce}
\end{equation}
\begin{equation}
\mathcal{L}_{\text{TRBL}}(y, \hat{y}) = \mathcal{L}_{\text{BCE}}(y, \hat{y}) \cdot (1 - y + \lambda y)
\label{eq:trbl_v1}
\end{equation}
where $\epsilon$ is a small constant added for numerical stability. The term $\left[ 1 + (\lambda - 1) \cdot y \right]$ ensures that only positive samples receive an increased weight of $\lambda$.

The total loss over a batch of $N$ samples is given by:
\begin{equation}
\mathcal{L}_{\text{total}} = \frac{1}{N} \sum_{i=1}^{N} \mathcal{L}_{\text{TRBL}}(y_i, \hat{y}_i)
\end{equation}
By setting $\lambda > 1$, the TRBL prioritizes sensitivity over specificity, which is a crucial design choice in TB triage applications where the cost of a false negative can be significantly higher than that of a false positive.

\begin{table*}[t!]
\centering
\small

\resizebox{\linewidth}{!}{
\begin{tabular}{lccccccc}
\toprule
\textbf{Model} & \textbf{Params} & \textbf{Accuracy $\uparrow$} & \textbf{True Positive Rate $\uparrow$} & \textbf{True Negative Rate $\uparrow$} & \textbf{F1-score $\uparrow$} & \textbf{AUROC $\uparrow$} & \textbf{TT(s) $\downarrow$} \\
\midrule
\textbf{Logistic} \cite{Cox1958} & $<$0.01M & 0.781 & 0.841 & 0.458 & 0.706 & 0.824 & 7.425 \\
\hspace{1em}\textit{- w/o Audio} & & 0.786 & 0.855 & 0.459 & 0.716 & 0.819 & \textbf{0.327} \\
\hspace{1em}\textit{- w/o Tabular} & & 0.735 & 0.895 & 0.299 & 0.686 & 0.633 & 5.530 \\
\midrule

\textbf{LightGBM} \cite{Ke2017} & 1.2M & 0.778 & 0.762 & 0.815 & 0.783 & 0.834 & 45.215\\

\hspace{1em}\textit{- w/o Audio} & & 0.814 & 0.847 & 0.825 & 0.838 & 0.859 & 11.450\\

\hspace{1em}\textit{- w/o Tabular} & & 0.728 & 0.898 & 0.423 & 0.730 & 0.693 & 27.359\\
\midrule
\textbf{1D-CNN} \cite{Kiranyaz2016} & 2.1M & 0.755 & 0.983 & 0.436 & 0.783 & 0.809 & 21.534 \\
\hspace{1em}\textit{- w/o Audio} & & 0.760 & 0.966 & 0.527 & 0.793 & 0.812 & 19.288\\
\hspace{1em}\textit{- w/o Tabular} & & 0.738 & 0.954 & 0.615 & 0.812 & 0.797 & 20.844\\
\midrule

\textbf{ResNet34} \cite{He2016} & 21.8M & 0.757 & 0.511 & 0.490 & 0.500 & 0.731 & 502.463\\

\hspace{1em}\textit{- w/o Audio} & & \ding{55} & \ding{55} & \ding{55} & \ding{55} & \ding{55} & \ding{55} \\

\hspace{1em}\textit{- w/o Tabular} & & 0.719 & 0.654 & 0.442 & 0.597 & 0.687 & 398.565\\
\midrule
\textbf{Wav2vec-2.0-Large} \cite{Baevski2020} & 317M & \ding{55} & \ding{55} & \ding{55} & \ding{55} & \ding{55} & \ding{55} \\
\hspace{1em}\textit{- w/o Audio} & & \ding{55} & \ding{55} & \ding{55} & \ding{55} & \ding{55} & \ding{55} \\
\hspace{1em}\textit{- w/o Tabular} & & 0.729 & 0.808 & 0.389 & 0.672 & 0.691 & 1898.435\\
\midrule

\textbf{TabTransformer} \cite{huang2020tabtransformer} & 8.9M & 0.733 & 0.785 & 0.617 & 0.721 & 0.737 & 121.558 \\

\hspace{1em}\textit{- w/o Audio} & & 0.721 & 0.760 & 0.591 & 0.715 & 0.718 & 44.778 \\

\hspace{1em}\textit{- w/o Tabular} & & 0.708 & 0.732 & 0.568 & 0.700 & 0.701 & 79.115\\
\midrule

\textbf{CNN-LightGBM Ensemble} 
\cite{Lu2023} & 2.8M & 0.788 & 0.812 & 0.692 & 0.768 & 0.792 & 31.225 \\
\hspace{1em}\textit{- w/o Audio} & & 0.728 & 0.715 & 0.586 & 0.733 & 0.780 & 5.570 \\
\hspace{1em}\textit{- w/o Tabular} & & 0.735 & 0.790 & 0.621 & 0.721 & 0.760 & 25.433\\
\midrule
\textbf{HeAR} \cite{hear2024} & - & \ding{55} & \ding{55} & \ding{55} & \ding{55} & \ding{55} & \ding{55} \\

\hspace{1em}\textit{- w/o Audio} & & \ding{55} & \ding{55} & \ding{55} & \ding{55} & \ding{55} & \ding{55} \\

\hspace{1em}\textit{- w/o Tabular} & & \underline{0.768} & \underline{0.862} & \underline{0.674} & \underline{0.807} & \underline{0.768} & \underline{\ding{55}}\\
\midrule
\textbf{Qwen-Omni 3B} \cite{Qwen2.5-Omni} & 3B & 0.812 & 0.895 & 0.855 & 0.845 & 0.900 & 4531.274 \\
\hspace{1em}\textit{- w/o Audio} & & 0.790 & 0.830 & 0.710 & 0.783 & 0.815 & 2815.631 \\
\hspace{1em}\textit{- w/o Tabular} & & 0.801 & 0.900 & 0.755 & 0.835 & 0.885 & 4158.993 \\
\midrule

\textbf{DeepGB-TB (Ours)} & 5.2M & \textbf{0.817} & \textbf{0.902} & \textbf{0.866} & \textbf{0.851} & \textbf{0.903} & 44.595\\

\hspace{1em}\textit{- w/o Audio} & & 0.785 & 0.770 & 0.820 & 0.790 & 0.840 & 22.383\\

\hspace{1em}\textit{- w/o Tabular} & & 0.771 & 0.898 & 0.701 & 0.818 & 0.825 & 27.588\\
\bottomrule
\end{tabular}
}
\caption{Comparative performance analysis of the proposed framework against SOTA and baseline models. \ding{55} indicates lack of compatibility for data type or convergence failure. The underline indicates zero-shot (as HeAR only supports API-based inference). TT(s) represent training time (seconds). "w/o" stands for "without". Best results are in bold.}
\label{tab1}
\end{table*}

\section{Experiments}

\begin{figure}[t]
\centering
\includegraphics[width=0.9\columnwidth]{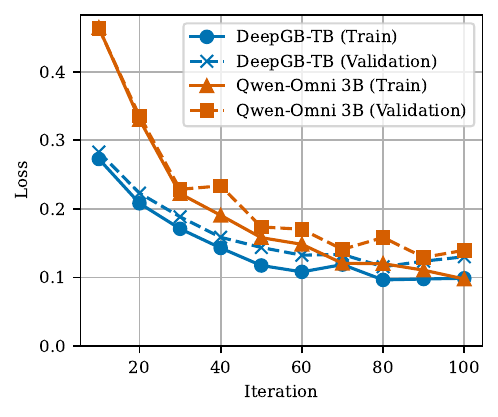} 
\caption{Comparison of Model Training and Validation Loss. The x-axis denotes epochs for DeepGB-TB and training steps for Qwen-Omni.}
\label{fig3}
\end{figure}

\subsection{Experimental Setup}
\label{subsec:setup}

We employed a 5-fold stratified cross-validation scheme for robust evaluation, which is ideal for the imbalanced nature of the dataset. All models were implemented in PyTorch 2.6.0 with CUDA 12.4 and Python 3.10, accelerated by an NVIDIA A100 80G GPU. We used the AdamW optimizer~\citep{loshchilov2019} with a batch size of 64 and a cosine annealing learning rate schedule (minimum value of $1 \times 10^{-6}$). Training durations were set by model type to ensure a fair comparison. Our proposed DeepGB-TB, along with other models trained from scratch (1D-CNN, ResNet, TabTransformer, and the CNN-LightGBM Ensemble), were trained for 100 epochs. Following standard fine-tuning protocols, the large pre-trained models were trained for limited epochs: Wav2vec-2.0 for 3 epochs and Qwen-Omni with LoRA~\citep{hu2021loralowrankadaptationlarge} for a single epoch.

\begin{table*}[t]  
  \centering  
  \small

  \begin{tabular}{lcccccc}  
  \toprule  
    
  \textbf{Excluded Feature} & \textbf{Accuracy $\uparrow$} & \textbf{Sensitivity $\uparrow$} & \textbf{Specificity $\uparrow$} & \textbf{F1-score $\uparrow$} & \textbf{AUROC $\uparrow$} & \textbf{p-value*} \\  
  \midrule  
  \ding{55} (Baseline) & $0.765 \pm 0.065$ & $0.944 \pm 0.053$ & $0.567 \pm 0.129$ & $0.837 \pm 0.080$ & $0.887 \pm 0.034$ & \ding{55} \\  
  Gender & $0.759 \pm 0.044$ & $0.806 \pm 0.068$ & $0.786 \pm 0.131$ & $0.796 \pm 0.093$ & $0.888 \pm 0.024$ & 0.82 \\  
  Hemoptysis & $0.764 \pm 0.041$ & $0.782 \pm 0.116$ & $0.761 \pm 0.161$ & $0.777 \pm 0.125$ & $0.885 \pm 0.034$ & 0.41 \\  
  Weight Loss & $0.770 \pm 0.027$ & $0.739 \pm 0.078$ & $0.622 \pm 0.193$ & $0.722 \pm 0.130$ & $0.881 \pm 0.027$ & \bfseries0.05 \\  
  Smoke & $0.779 \pm 0.028$ & $0.819 \pm 0.059$ & $0.766 \pm 0.081$ & $0.805 \pm 0.072$ & $0.888 \pm 0.031$ & 0.87 \\  
  Fever & $0.755 \pm 0.039$ & $0.856 \pm 0.052$ & $0.677 \pm 0.040$ & $0.807 \pm 0.044$ & $0.888 \pm 0.035$ & 0.91 \\  
  Night Sweats & $0.779 \pm 0.020$ & $0.806 \pm 0.053$ & $0.792 \pm 0.125$ & $0.798 \pm 0.083$ & $0.894 \pm 0.032$ & 0.21 \\  
  Age & $0.753 \pm 0.029$ & $0.836 \pm 0.083$ & $0.729 \pm 0.078$ & $0.807 \pm 0.072$ & $0.891 \pm 0.031$ & 0.57 \\  
  Height & $0.776 \pm 0.014$ & $0.748 \pm 0.158$ & $0.741 \pm 0.084$ & $0.754 \pm 0.124$ & $0.888 \pm 0.031$ & 0.87 \\  
  Weight & $0.759 \pm 0.030$ & $0.843 \pm 0.086$ & $0.525 \pm 0.266$ & $0.780 \pm 0.157$ & $0.886 \pm 0.035$ & 0.56 \\  
  Cough Duration & $0.754 \pm 0.046$ & $0.833 \pm 0.056$ & $0.735 \pm 0.065$ & $0.807 \pm 0.063$ & $0.883 \pm 0.034$ & 0.21 \\  
  Heart Rate & $0.764 \pm 0.013$ & $0.776 \pm 0.157$ & $0.724 \pm 0.197$ & $0.767 \pm 0.154$ & $0.882 \pm 0.029$ & \bfseries0.01 \\  
  Temperature & $0.764 \pm 0.031$ & $0.802 \pm 0.103$ & $0.666 \pm 0.092$ & $0.776 \pm 0.093$ & $0.885 \pm 0.038$ & 0.45 \\  
  ZCR & $0.790 \pm 0.029$ & $0.773 \pm 0.059$ & $0.804 \pm 0.160$ & $0.783 \pm 0.108$ & $0.890 \pm 0.029$ & 0.74 \\  
  Centroid & $0.800 \pm 0.029$ & $0.756 \pm 0.118$ & $0.679 \pm 0.048$ & $0.745 \pm 0.089$ & $0.886 \pm 0.031$ & 0.52 \\  
  F0 & $0.762 \pm 0.018$ & $0.793 \pm 0.061$ & $0.607 \pm 0.188$ & $0.751 \pm 0.116$ & $0.889 \pm 0.034$ & 0.92 \\  
  Energy & $0.798 \pm 0.030$ & $0.740 \pm 0.114$ & $0.552 \pm 0.279$ & $0.707 \pm 0.160$ & $0.885 \pm 0.036$ & 0.43 \\  
  Chroma Vector & $0.792 \pm 0.025$ & $0.705 \pm 0.088$ & $0.581 \pm 0.280$ & $0.687 \pm 0.165$ & $0.890 \pm 0.029$ & 0.74 \\  
  MFCCs & $0.787 \pm 0.027$ & $0.803 \pm 0.074$ & $0.671 \pm 0.348$ & $0.777 \pm 0.182$ & $0.879 \pm 0.041$ & \bfseries0.05 \\  
  Mel-Spectrogram & $0.802 \pm 0.035$ & $0.763 \pm 0.046$ & $0.870 \pm 0.116$ & $0.790 \pm 0.099$ & $0.902 \pm 0.024$ & \bfseries$<$0.01 \\  
        \bottomrule
    \end{tabular}
    \caption{Model performance comparison of DeepGB-TB with excluded features. * indicate independent samples t-test for AUROC compared to the full-variable (baseline); Data are mean $\pm$SD. Bold values indicate statistical significance (p-value$\leq$0.05).}

\label{tab:excluded_features}  
\end{table*}

\subsection{Comparative Analysis}
\label{subsec:analysis}
We evaluated all models using a comprehensive set of metrics: Accuracy, True Positive Rate (TPR, or Sensitivity), True Negative Rate (TNR, or Specificity), F1-score, and Area Under the Receiver Operating Characteristic Curve (AUROC). As detailed in Table~\ref{tab1}, our proposed DeepGB-TB achieves SOTA performance (0.903 AUROC, 0.851 F1-score). This surpasses unimodal baselines like TabTransformer (0.737 AUROC) and audio-only foundation models like HeAR (0.768 AUROC), as well as a late-fusion CNN-LightGBM Ensemble (0.792 AUROC), highlighting our integrated architecture's superiority. Notably, our model achieves superior multimodal performance over the larger Qwen-Omni 3B (0.903 vs. 0.900 AUROC), despite the latter's strong audio-only result (0.885 AUROC). The framework's effective synergy is confirmed by internal ablations: the audio-only (0.825 AUROC) and tabular-only (0.840 AUROC) versions perform significantly worse than the fully integrated model, demonstrating our framework's effective synergy between cough acoustics and demographic data to enhance TB diagnosis, and its robust predictive performance reinforces its clinical reliability.

\begin{table}[t!]
\centering

\begin{tabular}{lcc}
\toprule
\textbf{Module Configuration} & \textbf{F1-score $\uparrow$} & \textbf{AUROC $\uparrow$} \\
\midrule
CNN-Backbone & 0.783 & 0.809 \\
\hspace{1em}+ CVPM & 0.795 \scriptsize(+1.5\%) & 0.822 \scriptsize(+1.6\%) \\
\hspace{1em}+ IMDM & 0.830 \scriptsize(+6.0\%) & 0.889 \scriptsize(+9.9\%) \\
\hspace{1em}+ CM-BCA & 0.845 \scriptsize(+7.9\%) & 0.901 \scriptsize(+11.4\%) \\
 \hspace{1em}+ TRBL & 0.851 \scriptsize(+8.7\%) & 0.903 \scriptsize(+11.6\%) \\
\midrule
\hspace{1em}w/o CM-BCA & 0.840 \scriptsize(-1.3\%) & 0.891 \scriptsize(-1.3\%) \\
\hspace{1em}w/o TRBL & 0.845 \scriptsize(-0.7\%) & 0.901 \scriptsize(-0.2\%) \\
\bfseries DeepGB-TB & \bfseries0.851 & \bfseries0.903 \\
\bottomrule
\end{tabular}
\caption{Ablation study of module contributions.}
\label{tab:ablation1}
\end{table}

\begin{table}
\centering

\begin{tabular}{l|c|c|c|c|c}
\toprule
\textbf{Metrics} & \multicolumn{5}{c}{\textbf{Tuberculosis Risk-Balanced Loss}} \\
\cmidrule(lr){2-6}
& \(\lambda = 1\) & \(\lambda = 2\) & \textbf{\(\lambda = 3\)} & \(\lambda = 4\) & \(\lambda = 5\) \\
\midrule
AUROC $\uparrow$       & 0.901 & 0.902 & \textbf{0.903} & 0.901 & 0.900 \\
F1-Score $\uparrow$  & 0.847 & 0.850 & \textbf{0.851} & 0.849 & 0.846 \\
\bottomrule
\end{tabular}
\caption{Ablation study of different \(\lambda\) values for TRBL.}
\label{tab:trbl}
\label{tab:ablation2}
\end{table}

\subsection{Ablation Studies}
\label{subsec:ablation}

We conducted a comprehensive ablation study to quantify the contribution of each module, with results in Table~\ref{tab:ablation1}. Starting from a CNN-Backbone baseline (0.809 AUROC), we incrementally added components: the CVPEM improved AUROC to 0.822 (+1.6\%), subsequently adding the IMDM substantially boosted it to 0.889 (+9.9\%), and the CM-BCA module further advanced it to 0.901 (+11.4\%). Applying the TRBL achieved peak performance (0.903 AUROC). This step-wise improvement demonstrates component synergy; conversely, removing modules like CM-BCA from the final model incurs a significant 1.3\% AUROC drop. We also optimized the TRBL hyperparameter $\lambda$, which weights the false negative penalty. As shown in Table~\ref{tab:trbl}, performance peaked at $\lambda=3$ (0.903 AUROC) when testing values from 1 to 5, confirming that a calibrated, risk-sensitive loss is critical for the clinical priority of minimizing missed TB cases. Finally, we evaluated the on-device inference of the FP16-quantized model using ONNX Runtime (Table~\ref{tab:deployment}). The model exhibits strong real-time capabilities, achieving a 142 ms latency on an iPhone 14 Pro while maintaining 0.903 AUROC, confirming its efficiency and feasibility for deployment on resource-constrained hardware.

\begin{figure}[t!]
\centering
\includegraphics[width=0.8\columnwidth]{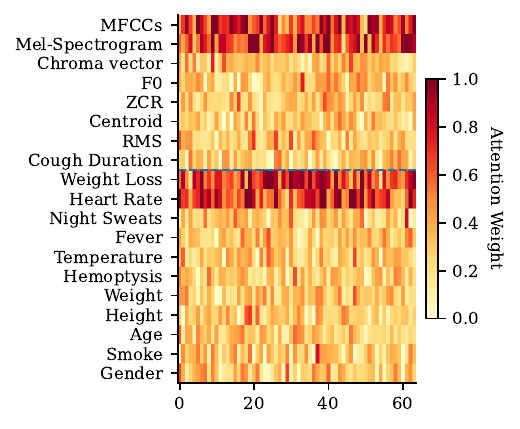} 

\vspace{-1em}
\caption{Attention heatmap over input features.}
\label{fig:attention}
\vspace{-1em}
\end{figure}

\begin{table}[t]\small          
\centering
\setlength{\tabcolsep}{6pt}     

\begin{tabular}{lccc}
\toprule
\multirow{2}{*}{\textbf{Device}} & \multicolumn{2}{c}{\textbf{Latency} $\downarrow$} &
\multirow{2}{*}{\textbf{AUROC} $\uparrow$} \\ \cmidrule(lr){2-3}
 & \textbf{Mean (ms)} & \textbf{Std (ms)} & \\
\midrule
Google Pixel 6 Pro   & 185 & 4.7  & 0.901 \\
Samsung Galaxy S22          & 210 & 6.3  & 0.902 \\
Apple iPhone 14 Pro         & 142 & 3.9  & 0.903 \\
Jetson Nano (MAXN)          & 278 & 8.6  & 0.899 \\
Raspberry Pi 4 (4GB)       & 450 & 12.1 & 0.898 \\
\bottomrule
\end{tabular}
\caption{On‑device inference performance of the quantized \textsc{DeepGB‑TB} model  
(FP16, ONNX Runtime). Results are averaged over 100 test samples.}
\label{tab:deployment}
\end{table}

\subsection{Feature Importance Analysis}
\label{subsec:feature_importance}

To evaluate feature importance, we performed a leave-one-out analysis (Table~\ref{tab:excluded_features}), iteratively training the model with one feature excluded and comparing its AUROC to the full-feature baseline. A statistically significant performance drop (p$<$0.05) indicates feature importance. Results show that both clinical and acoustic features are vital. Excluding Mel-Spectrograms caused the most significant degradation (p$<$0.01), underscoring the critical role of spectral representation. Among clinical data, excluding Hemoptysis (p=0.05), Weight Loss (p=0.05), or Heart Rate (p=0.01) significantly reduced performance, as did excluding MFCCs (p=0.05). We further visualize feature importance via attention weights in Figure~\ref{fig:attention}, where color intensity reflects magnitude (mean-computed for high-dimensional features) and the x-axis denotes batch samples. The model consistently assigns higher weights to key audio features (MFCCs, Mel-Spectrogram) and clinical signals (Heart Rate, Weight Loss), demonstrating its ability to prioritize discriminative inputs. Both analyses validate that our multimodal model effectively leverages a combination of high-level acoustic patterns and key clinical indicators for its predictions.


\section{Conclusion and Limitations}
\label{sec:conclusion}
This paper introduces DeepGB-TB, a novel and lightweight multimodal framework for TB screening. The framework deeply and efficiently fuses cough audio and demographic data via proposed CVPEM and IMDM hybrid architecture, which facilitates interaction between the custom-designed CM-BCA. In contrast to computationally expensive LMMs or simplistic late-fusion methods, our specialized architecture achieves SOTA performance on a multicenter dataset. Despite achieving exceptional performance, our interpretable and efficient open-source model’s validation on retrospective data necessitates future prospective trials to confirm its generalizability and real-world impact.

\section{Acknowledgments}
The datasets used for the analyses described were contributed by Dr. Adithya Cattamanchi at UCSF and Dr. Simon Grandjean Lapierre at University of Montreal and were generated in collaboration with researchers at Stellenbosch University (PI Grant Theron), Walimu (PIs William Worodria and Alfred Andama); De La Salle Medical and Health Sciences Institute (PI Charles Yu), Vietnam National Tuberculosis Program (PI Nguyen Viet Nhung), Christian Medical College (PI DJ Christopher), Centre Infectiologie Charles Mérieux Madagascar (PIs Mihaja Raberahona \& Rivonirina Rakotoarivelo), and Ifakara Health Institute (PIs Issa Lyimo \& Omar Lweno) with funding from the U.S. National Institutes of Health (U01 AI152087), The Patrick J. McGovern Foundation and Global Health Labs.

\bibliography{aaai2026}

\begin{thebibliography}{38}
\providecommand{\natexlab}[1]{#1}

\bibitem[{Al-Zoghby et~al.(2025)Al-Zoghby, Ismail~Ebada, Saleh, Abdelhay, and Awad}]{multimodal_review}
Al-Zoghby, A.~M.; Ismail~Ebada, A.; Saleh, A.~S.; Abdelhay, M.; and Awad, W.~A. 2025.
\newblock A comprehensive review of multimodal deep learning for enhanced medical diagnostics.
\newblock \emph{Computers, Materials \& Continua}, 84(3).

\bibitem[{Babyak(2004)}]{Babyak2001}
Babyak, M.~A. 2004.
\newblock What you see may not be what you get: a brief, nontechnical introduction to overfitting in regression-type models.
\newblock \emph{Biopsychosocial Science and Medicine}, 66(3): 411--421.

\bibitem[{Baevski et~al.(2020)Baevski, Zhou, Mohamed, and Auli}]{Baevski2020}
Baevski, A.; Zhou, H.; Mohamed, A.; and Auli, M. 2020.
\newblock wav2vec 2.0: A Framework for Self-Supervised Learning of Speech Representations.
\newblock In \emph{Advances in Neural Information Processing Systems 33}, 12449--12460.

\bibitem[{Bagcchi(2023)}]{who2023}
Bagcchi, S. 2023.
\newblock WHO's global tuberculosis report 2022.
\newblock \emph{The Lancet Microbe}, 4(1): e20.

\bibitem[{Baur et~al.(2024)Baur, Nabulsi, Weng, Garrison, Blankemeier, Fishman, Chen, Kakarmath, Maimbolwa, Sanjase et~al.}]{hear2024}
Baur, S.; Nabulsi, Z.; Weng, W.-H.; Garrison, J.; Blankemeier, L.; Fishman, S.; Chen, C.; Kakarmath, S.; Maimbolwa, M.; Sanjase, N.; et~al. 2024.
\newblock HeAR--Health Acoustic Representations.
\newblock \emph{arXiv preprint arXiv:2403.02522}.

\bibitem[{Boehme et~al.(2010)Boehme, Nabeta, Hillemann et~al.}]{boehme2010}
Boehme, P.; Nabeta, P.; Hillemann, D.; et~al. 2010.
\newblock Rapid molecular detection of tuberculosis and rifampicin resistance.
\newblock \emph{N. Engl. J. Med.}, 363(11): 1005--1015.

\bibitem[{Boll(1979)}]{ref_noise1}
Boll, S.~F. 1979.
\newblock Suppression of acoustic noise in speech using spectral subtraction.
\newblock \emph{IEEE Trans. Acoust. Speech Signal Process.}, 27(2): 113--120.

\bibitem[{Chakravorty et~al.(2017)Chakravorty, Simmons, Rowneki, Parmar, Cao, Ryan, Banada, Deshpande, Shenai, Gall et~al.}]{chakravorty2017}
Chakravorty, S.; Simmons, A.~M.; Rowneki, M.; Parmar, H.; Cao, Y.; Ryan, J.; Banada, P.~P.; Deshpande, S.; Shenai, S.; Gall, A.; et~al. 2017.
\newblock The new Xpert MTB/RIF Ultra: improving detection of Mycobacterium tuberculosis and resistance to rifampin in an assay suitable for point-of-care testing.
\newblock \emph{MBio}, 8(4): 10--1128.

\bibitem[{Chen et~al.(2025)Chen, Wang, Du, Sun, Wang, Ni, An, Fan, Li, Guo et~al.}]{who2024}
Chen, Z.; Wang, T.; Du, J.; Sun, L.; Wang, G.; Ni, R.; An, Y.; Fan, X.; Li, Y.; Guo, R.; et~al. 2025.
\newblock Decoding the WHO global tuberculosis report 2024: a critical analysis of global and Chinese key data.
\newblock \emph{Zoonoses}, 5(1): 999.

\bibitem[{Cox(1958)}]{Cox1958}
Cox, D. 1958.
\newblock The Regression Analysis of Binary Sequences.
\newblock \emph{Journal of the Royal Statistical Society: Series B (Methodological)}, 20(2): 215--242.

\bibitem[{Davis and Mermelstein(1980)}]{davis1980}
Davis, S.; and Mermelstein, P. 1980.
\newblock Comparison of parametric representations for monosyllabic word recognition in continuously spoken sentences.
\newblock In \emph{IEEE Transactions on Acoustics, Speech, and Signal Processing}, volume~28, 357--366.

\bibitem[{Fawcett(2006)}]{Fawcett2006}
Fawcett, T. 2006.
\newblock An introduction to ROC analysis.
\newblock \emph{Pattern recognition letters}, 27(8): 861--874.

\bibitem[{Giannakopoulos and Pikrakis(2014)}]{giannakopoulos2015}
Giannakopoulos, T.; and Pikrakis, A. 2014.
\newblock \emph{Introduction to Audio Analysis: A MATLAB Approach}.
\newblock Academic Press.

\bibitem[{Goodfellow et~al.(2016)Goodfellow, Bengio, Courville, and Bengio}]{crossentropy}
Goodfellow, I.; Bengio, Y.; Courville, A.; and Bengio, Y. 2016.
\newblock \emph{Deep learning}, volume~1.
\newblock MIT press Cambridge.

\bibitem[{Hastie, Tibshirani, and Friedman(2009)}]{stat_method2}
Hastie, T.; Tibshirani, R.; and Friedman, J. 2009.
\newblock \emph{The Elements of Statistical Learning}.
\newblock New York: Springer, 2nd edition.

\bibitem[{He et~al.(2016)He, Zhang, Ren, and Sun}]{He2016}
He, K.; Zhang, X.; Ren, S.; and Sun, J. 2016.
\newblock Deep Residual Learning for Image Recognition.
\newblock In \emph{Proceedings of the IEEE Conference on Computer Vision and Pattern Recognition (CVPR)}, 770--778.

\bibitem[{Hu et~al.(2021)Hu, Shen, Wallis, Allen-Zhu, Li, Wang, Wang, and Chen}]{hu2021loralowrankadaptationlarge}
Hu, E.~J.; Shen, Y.; Wallis, P.; Allen-Zhu, Z.; Li, Y.; Wang, S.; Wang, L.; and Chen, W. 2021.
\newblock LoRA: Low-Rank Adaptation of Large Language Models.
\newblock arXiv:2106.09685.

\bibitem[{Huang et~al.(2020)Huang, Khetan, Cvitkovic, and Karnin}]{huang2020tabtransformer}
Huang, X.; Khetan, A.; Cvitkovic, M.; and Karnin, Z. 2020.
\newblock TabTransformer: Tabular Data Modeling Using Contextual Embeddings.
\newblock arXiv:2012.06678.

\bibitem[{Jabeen et~al.(2023)Jabeen, Li, Amin, Bourahla, Li, and Jabbar}]{ramachandram2017}
Jabeen, S.; Li, X.; Amin, M.~S.; Bourahla, O.; Li, S.; and Jabbar, A. 2023.
\newblock A review on methods and applications in multimodal deep learning.
\newblock \emph{ACM Transactions on Multimedia Computing, Communications and Applications}, 19(2s): 1--41.

\bibitem[{Jaganath et~al.(2024)Jaganath, Sieberts, Raberahona, Huddart, Omberg, Rakotoarivelo, Lyimo, Lweno, Christopher, Nhung, Worodria, Yu, Chen, Chen, Chen, Huang, Huang, Mulier, Rafter, Shih, Tsao, Wang, Wu, Bachman, Burkot, Dewan, Kulhare, Small, Yadav, Grandjean~Lapierre, Theron, and Cattamanchi}]{jaganath2024}
Jaganath, D.; Sieberts, S.~K.; Raberahona, M.; Huddart, S.; Omberg, L.; Rakotoarivelo, R.; Lyimo, I.; Lweno, O.; Christopher, D.~J.; Nhung, N.~V.; Worodria, W.; Yu, C.; Chen, J.-Y.; Chen, S.-H.; Chen, T.-M.; Huang, C.-H.; Huang, K.-L.; Mulier, F.; Rafter, D.; Shih, E. S.~C.; Tsao, Y.; Wang, H.-K.; Wu, C.-H.; Bachman, C.; Burkot, S.; Dewan, P.; Kulhare, S.; Small, P.~M.; Yadav, V.; Grandjean~Lapierre, S.; Theron, G.; and Cattamanchi, A. 2024.
\newblock Accelerating cough-based algorithms for pulmonary tuberculosis screening: Results from the CODA TB DREAM Challenge.
\newblock \emph{medRxiv}.

\bibitem[{Kant and Srivastava(2018)}]{imran2019}
Kant, S.; and Srivastava, M.~M. 2018.
\newblock Towards automated tuberculosis detection using deep learning.
\newblock In \emph{2018 IEEE Symposium Series on Computational Intelligence (SSCI)}, 1250--1253. IEEE.

\bibitem[{Ke et~al.(2017)Ke, Meng, Finley, Wang, Chen, Ma, Ye, and Liu}]{Ke2017}
Ke, G.; Meng, Q.; Finley, T.; Wang, T.; Chen, W.; Ma, W.; Ye, Q.; and Liu, T.-Y. 2017.
\newblock LightGBM: A Highly Efficient Gradient Boosting Decision Tree.
\newblock In \emph{Advances in Neural Information Processing Systems 30}, 3146--3154.

\bibitem[{Kent(1985)}]{kent1985}
Kent, P. 1985.
\newblock \emph{Public Health Mycobacteriology: A Guide for the Level III Laboratory}.
\newblock U.S. Department of Health and Human Services, CDC.

\bibitem[{Kiranyaz, Ince, and Gabbouj(2016)}]{Kiranyaz2016}
Kiranyaz, S.; Ince, T.; and Gabbouj, M. 2016.
\newblock Real-time patient-specific ECG classification by 1-D convolutional neural networks.
\newblock \emph{IEEE Trans. Biomed. Eng.}, 63(3): 664--675.

\bibitem[{Kuhn and Johnson(2013)}]{stat_method3}
Kuhn, M.; and Johnson, K. 2013.
\newblock \emph{Applied Predictive Modeling}.
\newblock New York: Springer.

\bibitem[{Lakhani and Sundaram(2017)}]{lakhani2017}
Lakhani, P.; and Sundaram, B. 2017.
\newblock Deep learning at chest radiography: automated classification of pulmonary tuberculosis by using convolutional neural networks.
\newblock \emph{Radiology}, 284(2): 574--582.

\bibitem[{Li et~al.(2025{\natexlab{a}})Li, Lu, Tang, Lai, Hu, Zhang, Xue, Wu, Razzak, Li et~al.}]{li2025rhythm}
Li, Y.; Lu, Z.; Tang, F.; Lai, S.; Hu, M.; Zhang, Y.; Xue, H.; Wu, Z.; Razzak, I.; Li, Q.; et~al. 2025{\natexlab{a}}.
\newblock Rhythm of Opinion: A Hawkes-Graph Framework for Dynamic Propagation Analysis.
\newblock \emph{arXiv preprint arXiv:2504.15072}.

\bibitem[{Li et~al.(2025{\natexlab{b}})Li, Zhang, Chen, Tang, Lu, Hu, Wu, Xue, Zhou, Li et~al.}]{li2025genesis}
Li, Y.; Zhang, Y.; Chen, R.; Tang, F.; Lu, Z.; Hu, M.; Wu, J.; Xue, H.; Zhou, M.; Li, C.; et~al. 2025{\natexlab{b}}.
\newblock Genesis: A Large-Scale Benchmark for Multimodal Large Language Model in Emotional Causality Analysis.
\newblock In \emph{Proceedings of the 33rd ACM International Conference on Multimedia}, 12651--12658.

\bibitem[{Linderman and Steinerberger(2019)}]{tsne_interpret}
Linderman, G.~C.; and Steinerberger, S. 2019.
\newblock Clustering with t-SNE, provably.
\newblock \emph{SIAM journal on mathematics of data science}, 1(2): 313--332.

\bibitem[{Loshchilov and Hutter(2019)}]{loshchilov2019}
Loshchilov, I.; and Hutter, F. 2019.
\newblock Decoupled Weight Decay Regularization.
\newblock In \emph{International Conference on Learning Representations (ICLR)}.

\bibitem[{Lu(2023)}]{Lu2023}
Lu, Z. 2023.
\newblock Deep Learning-based Decision-tree Classifier for Tuberculosis Diagnosis.
\newblock In \emph{2023 5th International Academic Exchange Conference on Science and Technology Innovation (IAECST)}, 1491--1495.

\bibitem[{Maaten and Hinton(2008)}]{tsne2008}
Maaten, L. v.~d.; and Hinton, G. 2008.
\newblock Visualizing data using t-SNE.
\newblock \emph{Journal of machine learning research}, 9(Nov): 2579--2605.

\bibitem[{Mantel(1967)}]{mantel1967}
Mantel, N. 1967.
\newblock The detection of disease clustering and a generalized regression approach.
\newblock \emph{Cancer Res.}, 27(2): 209--220.

\bibitem[{Sathishkumar et~al.(2024)Sathishkumar, Karunamurthy, Saint~Jesudoss, Anusri, Ugamaalya, and Valarmathy}]{prev_best}
Sathishkumar, R.; Karunamurthy, A.; Saint~Jesudoss, S.; Anusri, E.; Ugamaalya, C.; and Valarmathy, S. 2024.
\newblock Multimodal Deep Learning for Precise Lung Pathology Discrimination.
\newblock \emph{2024 International Conference on System, Computation, Automation and Networking (ICSCAN)}, 1--6.

\bibitem[{Snider~Jr and Roper(1992)}]{lawn2011}
Snider~Jr, D.~E.; and Roper, W.~L. 1992.
\newblock The new tuberculosis.
\newblock \emph{New England Journal of Medicine}, 326(10): 703--705.

\bibitem[{Uplekar et~al.(2015)Uplekar, Weil, Lonnroth, Jaramillo, Lienhardt, Dias, Falzon, Floyd, Gargioni, Getahun et~al.}]{who2015}
Uplekar, M.; Weil, D.; Lonnroth, K.; Jaramillo, E.; Lienhardt, C.; Dias, H.~M.; Falzon, D.; Floyd, K.; Gargioni, G.; Getahun, H.; et~al. 2015.
\newblock WHO's new end TB strategy.
\newblock \emph{The Lancet}, 385(9979): 1799--1801.

\bibitem[{Xu et~al.(2025)Xu, Guo, He, Hu, He, Bai, Chen, Wang, Fan, Dang, Zhang, Wang, Chu, and Lin}]{Qwen2.5-Omni}
Xu, J.; Guo, Z.; He, J.; Hu, H.; He, T.; Bai, S.; Chen, K.; Wang, J.; Fan, Y.; Dang, K.; Zhang, B.; Wang, X.; Chu, Y.; and Lin, J. 2025.
\newblock Qwen2.5-Omni Technical Report.
\newblock \emph{arXiv preprint arXiv:2503.20215}.

\bibitem[{Zhang et~al.(2025)Zhang, Li, Tang, Yu, Hu, Lu, Xue, Wu, Dang, Razzak, and Su}]{LiAVSS}
Zhang, Y.; Li, Y.; Tang, F.; Yu, Z.; Hu, M.; Lu, Z.; Xue, H.; Wu, Z.; Dang, K.; Razzak, I.; and Su, J. 2025.
\newblock Decoding the Flow: CauseMotion for Emotional Causality Analysis in Long-form Conversations.
\newblock In \emph{2025 IEEE International Conference on Advanced Visual and Signal-Based Systems (AVSS)}, 1--6.

\end{thebibliography}

\end{document}